\documentclass[letterpaper, 10 pt, conference]{ieeeconf}
\IEEEoverridecommandlockouts
\usepackage{cite}
\usepackage{amsmath,amssymb,amsfonts}
\usepackage{graphicx}
\usepackage{booktabs}
\usepackage{multirow}
\usepackage{array}
\usepackage{xcolor}
\usepackage{colortbl}
\usepackage{tikz}
\usetikzlibrary{arrows.meta,positioning,fit,calc}
\usepackage{placeins}
\usepackage[final]{microtype}
\usepackage{balance}
\usepackage[hidelinks]{hyperref}
\hypersetup{
  pdftitle={Mesh-Free Keypoint Learning for Category-Level 6D Object Pose Estimation},
  pdfauthor={Junghoon Seo, Hakjin Lee, Jaehoon Sim}
}

\newcommand{\R}{\mathbb{R}}
\newcommand{\Pobs}{\mathcal{P}}
\newcommand{\Mmesh}{\mathcal{M}}
\newcommand{\Rgt}{\mathbf{R}}
\newcommand{\tgt}{\mathbf{t}}
\newcommand{\sgt}{\mathbf{s}}
\newcommand{\pvec}{\mathbf{p}}
\newcommand{\qvec}{\mathbf{q}}
\newcommand{\Omesh}{\Omega_{\mathrm{mesh}}}
\newcommand{\Oenv}{\Omega_{\mathrm{env}}}
\newcommand{\degtwo}{$5^\circ2\,\mathrm{cm}$}
\newcommand{\degfive}{$5^\circ5\,\mathrm{cm}$}
\newcommand{\tendegtwo}{$10^\circ2\,\mathrm{cm}$}
\newcommand{\tendegfive}{$10^\circ5\,\mathrm{cm}$}

\definecolor{cpBlue}{HTML}{0072B2}
\definecolor{cpOrange}{HTML}{D55E00}
\definecolor{cpGreen}{HTML}{009E73}

\definecolor{cpOursRow}{gray}{0.90}
\newcommand{\oursrow}{\rowcolor{cpOursRow}}

\newcommand{\supbox}[1]{\makebox[6.6pt][l]{#1}}
\newcommand{\meshsup}{%
  \supbox{\textcolor{black!75}{\rule[-0.3pt]{3.7pt}{3.7pt}}}}
\newcommand{\labelsup}{%
  \supbox{\raisebox{0.15pt}{\begingroup
    \setlength{\fboxsep}{0pt}\setlength{\fboxrule}{0.5pt}%
    \textcolor{black!55}{\framebox[3.7pt]{\rule{0pt}{2.7pt}}}%
  \endgroup}}}
\newcommand{\nosup}{\supbox{}}
\begin{document}

\title{\LARGE\bfseries Mesh-Free Keypoint Learning\\for Category-Level 6D Object Pose Estimation}

\author{%
  \authorblockN{Junghoon Seo\textsuperscript{*}, Hakjin Lee, and Jaehoon Sim%
    \thanks{\textsuperscript{*}Corresponding author: Junghoon Seo
      (\href{mailto:sjh@pitin-ev.com}{\texttt{sjh@pitin-ev.com}}).}}
  \authorblockA{PIT IN Corp., South Korea}}

\maketitle

\begin{abstract}
Category-level 6D pose estimation relates partial RGB-D observations of unseen
instances to a canonical frame shared across a category. Object-aware
reconstruction helps adaptive keypoints retain the geometry of the visible
surface. Common reconstruction targets select observed points by their
distance to a registered instance surface. We present MORSE-Pose, which
derives this support directly from standard pose and metric-size annotations.
We map observed 3D points to normalized object coordinates and retain those
inside a tolerance-dilated pose-and-size envelope. This construction is
consistent across camera frames and metric
scales. When the annotated extent bounds the object surface, the envelope
conservatively contains the support of a conventional surface-distance
selector. We integrate this target with multi-level RGB-D features, adaptive
keypoints, and learnable category context. Experiments on REAL275 and
HouseCat6D show that the envelope closely matches surface-based supervision at
both point and task levels, remains stable under boundary perturbations and
viewpoint changes, and supports competitive category-level pose accuracy across
benchmarks.
\end{abstract}

\section{Introduction}
\label{sec:intro}

Robotic manipulation requires a stable object-centered reference frame for
reasoning about grasps, placements, contacts, and tool motion. Instance-level
pose estimation obtains this frame by aligning an observation to a known object
model~\cite{posecnn,densefusion}. Category-level 6D pose estimation addresses a
broader setting by predicting rotation, translation, and metric size for an
unseen instance of a known category in a shared canonical
frame~\cite{nocs,spd,fsnet}. A successful estimator
must establish the same semantic axes across substantial variation in shape,
appearance, viewpoint, and visibility. This capability allows a robot to
transfer an interaction strategy to a novel mug, bottle, or tool without first
acquiring its exact geometry.

\begin{figure}[t]
\centering
\includegraphics[width=\linewidth]{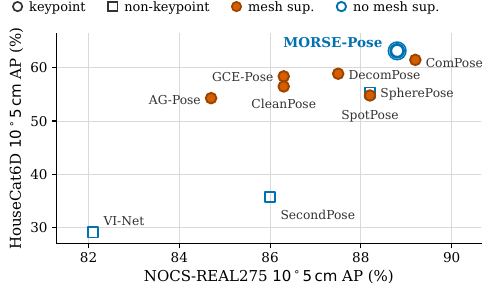}
\caption{Joint \tendegfive{} AP on REAL275 and HouseCat6D. Upper right is
better. Marker style indicates whether training uses registered instance
geometry.}
\label{fig:benchmark-teaser}
\end{figure}

Category-level pose estimation is both a geometric estimation problem and a
representation learning problem. An RGB-D crop records the object surface
visible from one viewpoint and can include points from supporting surfaces
or the background. Organizing the visible geometry in a category-consistent
frame helps relate corresponding parts across instances. Adaptive 3D
keypoints provide an effective interface for this
purpose~\cite{agpose,cleanpose,gcepose,compose}, aggregating dense observations
into a compact, instance-conditioned structure that supports
correspondence-based reasoning across intra-category shape variation.

Object-aware reconstruction provides complementary supervision by encouraging
this intermediate representation to retain the measured structure from which
pose is inferred, and its target determines which points in the RGB-D crop are
treated as object evidence. A registered instance mesh provides a precise
reference for selecting points near the object surface. Standard annotations
also describe where to select points for reconstruction. Pose defines the
canonical frame, and metric size gives the extent along each axis.

We use these annotations to construct a pose-and-size envelope. Each
observed 3D point is mapped to normalized object coordinates using the
annotated pose and size, and it contributes to reconstruction when it lies
inside the annotated extent after each face is dilated by an object-normalized
tolerance. The support follows the orientation and
anisotropic extent of the instance, and canonicalization gives the tolerance a
consistent meaning across camera frames, categories, and metric scales. When
the annotated extent bounds the object
surface, the dilated envelope contains every point a surface-distance
criterion selects, extending support to nearby measurements around concavities
and interiors.

MORSE-Pose, short for Mesh-free Object-aware Reconstruction via
pose-and-Size Envelopes, integrates this supervision with an adaptive RGB-D
keypoint estimator in which multi-level visual features, geometric evidence,
and learnable category prototypes shape keypoints that support correspondence
prediction, pose and size estimation, and observed-surface reconstruction,
while inference uses only the RGB-D crop and category
label. Our evaluation connects the supervision geometry to
the final task through point-level audits, matched target substitutions,
robustness and cross-view studies, ablations, and efficiency comparisons.
At the \tendegfive{} threshold in Fig.~\ref{fig:benchmark-teaser},
MORSE-Pose reaches $88.8$ AP on REAL275 and $63.2$ AP on HouseCat6D.

Our contributions are threefold: a pose-and-size envelope with containment and
invariance properties that derives object-aware reconstruction support
directly from category-level pose and metric-size annotations, its integration
with multi-level RGB-D features, adaptive keypoints, and learned category
context in a unified estimator, and an evaluation connecting selector geometry
to task behavior and strong pose accuracy on REAL275 and HouseCat6D.

\section{Related Work}
\label{sec:related}

Table~\ref{tab:requirements} summarizes the supervision required by common
geometry-aware reconstruction targets and positions the proposed envelope within
this landscape.

\subsection{Category-Level 6D Pose Estimation}
Pose estimators use different geometric references at inference. Instance-level
methods align observations to a known object model~\cite{posecnn,densefusion}.
Generalizable methods use reference views or a mesh of a novel
object~\cite{foundationpose,onepose}. NOCS~\cite{nocs} established the
category-level setting by mapping observations of unseen instances into a
shared canonical space to estimate pose and metric size jointly.

\begin{table}[t]
\centering
\caption{Supervision required by geometry-aware reconstruction targets.}
\label{tab:requirements}
\footnotesize
\setlength{\tabcolsep}{1.5pt}
\begin{tabular}{lccccc}
\toprule
Target & Mesh & Align. & Pose/size & Construction & Example \\
\midrule
Dense NOCS & Yes & Yes & Yes & Rendering & \cite{nocs,dpdn} \\
CAD completion & Yes & Yes & Yes & Surface match & \cite{spd,sgpa,compose} \\
Surface-gated rec. & Yes & Yes & Yes & Surface dist. & \cite{agpose,cleanpose,mempose} \\
Envelope (ours) & No & No & Yes & Box containment & -- \\
\bottomrule
\end{tabular}
\end{table}

Methods differ in how they connect an observation to this canonical space.
Shape-prior approaches deform a learned category template toward the observed
instance~\cite{spd,sgpa,dpdn,gcepose}, providing complete category-level shape
context. Observation-driven approaches infer canonical structure directly from
RGB-D through decoupled rotation estimation~\cite{fsnet,vinet}, geometry-guided
voting~\cite{gpvpose}, hybrid-scope point features~\cite{hspose}, implicit
canonical transformations~\cite{istnet}, sparse pose queries~\cite{query6dof},
transformer aggregation~\cite{catformer}, SE(3)-consistent RGB-D
fusion~\cite{secondpose}, and spherical or correspondence-centric
representations~\cite{spherepose,spotpose}. MORSE-Pose follows the
observation-driven setting and focuses on how object-aware reconstruction can be
supervised from category-level pose annotations.

\subsection{Adaptive Keypoints and Shape-Aware Learning}
Sparse keypoints provide a compact bridge between observed and canonical
geometry. AG-Pose~\cite{agpose} learns instance-adaptive keypoints from point
clouds and enriches them with local and global geometric relations. Subsequent
work incorporates causal feature adjustment and multimodal
distillation~\cite{cleanpose}, additional sensing modalities~\cite{mkpose},
stronger category context~\cite{gcepose}, category-specific
optimization~\cite{decompose}, progressive shape completion~\cite{compose}, and
memory over previously observed instances~\cite{mempose}.

Reconstruction and completion objectives provide complementary geometric
supervision. Dense canonical rendering and complete-shape prediction use
geometric assets to define targets on an object surface. For object-aware
reconstruction, the target consists of measured points in the RGB-D crop.
MORSE-Pose uses pose and metric size to select the points that guide the
keypoint representation.

\begin{figure*}[t]
\centering
\definecolor{pipeSlate}{HTML}{3B4856}
\colorlet{pipeTok}{cpBlue!55!cpGreen}
\newcommand{\pipesheet}[6]{%
  \begin{scope}[shift={(#1,#2)}, yslant=.3, xscale=.9]
    \fill[black, opacity=.16] (.55,-.55) rectangle ++(#3,#4);
    \node[anchor=south west, inner sep=0, outer sep=0, transform shape, #6] at (0,0)
      {\setkeys{Gin}{width=#3mm,height=#4mm,interpolate=true}#5};
    \draw[pipeSlate!65, line width=.3pt] (0,0) rectangle (#3,#4);
  \end{scope}}
\newcommand{\pipetokens}[3]{%
  \foreach \i/\v in {0/38, 1/78, 2/52, 3/92, 4/45, 5/66}
    \filldraw[fill=#3!\v, draw=white, line width=.45pt, rounded corners=.35pt]
      ({#1-5.375+1.85*\i}, {#2-.75}) rectangle ++(1.5,1.5);}
\begin{tikzpicture}[
  x=1mm, y=1mm, font=\footnotesize, line cap=round, line join=round,
  flow/.style={-{Latex[length=1.8mm,width=1.35mm]}, line width=.75pt, draw=pipeSlate},
  sup/.style={-{Latex[length=1.8mm,width=1.35mm]}, line width=.75pt, draw=cpOrange,
              dash pattern=on 2.8pt off 1.1pt on .5pt off 1.1pt},
  lab/.style={midway, above, inner sep=1.4pt},
  title/.style={anchor=base, inner sep=0, text=pipeSlate},
  card/.style={fill=white, draw=pipeSlate!40, line width=.4pt, rounded corners=1.6pt},
  shadow/.style={fill=black, opacity=.13, rounded corners=1.6pt},
  snowflake/.pic={
    \foreach \a in {0,60,120}{\draw (\a:1.15) -- (\a+180:1.15);}
    \foreach \a in {0,60,...,300}{
      \draw (\a:1.15) -- ++(\a+150:.45); \draw (\a:1.15) -- ++(\a-150:.45);}},
  nomesh/.pic={
    \fill[white] (0,0) circle (2.45);
    \draw[black!60, line width=.35pt]
      (-1.5,-1.5) grid[step=1.5] (1.5,1.5)
      (-1.5,0) -- (0,1.5) (-1.5,-1.5) -- (1.5,1.5) (0,-1.5) -- (1.5,0);
    \draw[cpOrange, line width=.8pt] (0,0) circle (2.45) (135:2.45) -- (-45:2.45);}
]
\begin{scope}
  \clip[rounded corners=2.4pt] (0,0) rectangle (177.8,-23);
  \shade[top color=pipeSlate!3, bottom color=pipeSlate!12] (0,0) rectangle (177.8,-23);
  \fill[pipeSlate] (0,0) rectangle (4,-23);
\end{scope}
\begin{scope}
  \clip[rounded corners=2.4pt] (0,-26.8) rectangle (177.8,-46.2);
  \fill[cpOrange!4] (0,-26.8) rectangle (177.8,-46.2);
  \draw[cpOrange!17, line width=.2pt, step=2] (4,-46.2) grid (177.8,-26.8);
  \fill[cpOrange] (0,-26.8) rectangle (4,-46.2);
\end{scope}
\node[rotate=90, text=white, font=\scriptsize\scshape] at (2,-11.5) {\textls[90]{inference}};
\node[rotate=90, text=white, font=\scriptsize\scshape] at (2,-36.5) {\textls[90]{training}};

\pipesheet{11.0}{-18.935}{11}{11}{\includegraphics{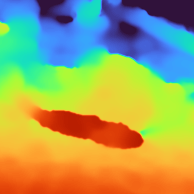}}{}
\pipesheet{6.0}{-21.435}{11}{11}{\includegraphics{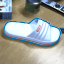}}{}
\node[title] at (13.45,-3.4) {RGB-D crop};
\draw[flow] (21.3,-13.2) -- (29.7,-13.2);
\shade[left color=cpGreen!72, right color=cpGreen!92!black, rounded corners=.9pt]
  (30,-20.2) -- (30,-6.2) -- (36.8,-8.7) -- (36.8,-17.7) -- cycle;
\node[rotate=90, text=white, font=\scriptsize\bfseries] at (33.7,-13.2) {DINOv2};
\filldraw[fill=white, draw=cpGreen!75!black, line width=.4pt] (30.5,-6.3) circle (1.7);
\pic[draw=cpGreen!75!black, line width=.42pt] at (30.5,-6.3) {snowflake};
\foreach \k/\feature in {3/{\includegraphics{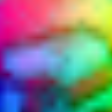}},
                         2/{\includegraphics{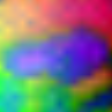}},
                         1/{\includegraphics{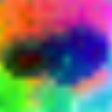}},
                         0/{\includegraphics{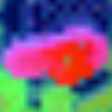}}}
  {\pipesheet{37.6+2.6*\k}{-21.529+1.15*\k}{10.4}{10.4}{\feature}{}}
\node[title] at (42.4,-3.4) {multi-level $\Phi_\ell$};
\draw[flow] (55.2,-13.2) -- node[lab]{$f_i$} (67.6,-13.2);
\fill[shadow] (68.3,-22.2) rectangle (87.5,-5.4);
\draw[card] (67.8,-21.7) rectangle (87.0,-4.9);
\node[inner sep=0] at (77.4,-13.3) {\includegraphics[height=15.8mm]{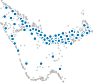}};
\node[title] at (77.4,-3.4) {keypoints $\mathbf{k}_j$};
\draw[flow] (87.2,-13.2) -- (103.8,-13.2);
\pipetokens{95.5}{-13.2}{cpBlue}
\node[above, inner sep=1.4pt] at (95.5,-12.45) {$\mathbf{g}_j$};
\pipesheet{108.8}{-18.154}{15.58}{8.6}{\includegraphics{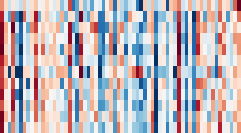}}{opacity=.3}
\pipesheet{106.4}{-19.604}{15.58}{8.6}{\includegraphics{figures/pipe_parts/context.png}}{opacity=.55}
\pipesheet{104.0}{-21.054}{15.58}{8.6}{\includegraphics{figures/pipe_parts/context.png}}{}
\node[title] at (113.4,-3.4) {category context $\mathbf{C}_c$};
\draw[flow] (123.3,-13.2) -- (140.8,-13.2);
\pipetokens{132.0}{-13.2}{pipeTok}
\node[above, inner sep=1.4pt] at (132.0,-12.45) {$\widetilde{\mathbf{g}}_j$};
\fill[shadow] (141.5,-22.2) rectangle (176.7,-5.4);
\draw[card] (141.0,-21.7) rectangle (176.2,-4.9);
\node[inner sep=0, anchor=west] at (141.6,-13.3) {\includegraphics[height=15.8mm]{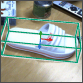}};
\draw[pipeSlate!25, line width=.3pt] (158.2,-6.2) -- (158.2,-20.4);
\node[inner sep=0, anchor=west] at (158.9,-13.3) {\includegraphics[height=15.8mm]{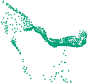}};
\node[title] at (158.6,-3.4) {pose, size, $\widehat{\Pobs}$};

\draw[sup] (10.6,-20.4) -- (10.6,-27.6);
\node[inner sep=0, anchor=west] at (6,-35.1) {\includegraphics[height=14.2mm]{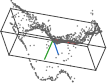}};
\node[title] at (16,-44.9) {points $+$ pose/size};
\pic at (30.6,-33) {nomesh};
\node[inner sep=0, text=pipeSlate] at (30.6,-38.7) {mesh-free};
\draw[sup] (37.4,-35.1) -- node[lab]{$\qvec_i$} (65.2,-35.1);
\fill[cpBlue, opacity=.2, rounded corners=2pt] (66.6,-43.1) rectangle (112.5,-28.1);
\filldraw[fill=white, draw=cpBlue, line width=.7pt, rounded corners=2pt]
  (66.0,-42.6) rectangle (111.9,-27.6);
\node[inner sep=0, anchor=west] at (66.9,-35.1) {\includegraphics[height=13.6mm]{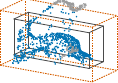}};
\node[inner sep=0] at (99.2,-34.6)
  {\small$\lvert q_{i,a}\rvert\le e_a/2+\textcolor{cpOrange}{\tau}$};
\node[fill=cpBlue, text=white, font=\footnotesize\bfseries, rounded corners=1.6pt,
      inner xsep=2.4pt, inner ysep=1.1pt] at (89.0,-43.9) {envelope $\boldsymbol{\Oenv}$};
\draw[sup] (112.7,-35.1) -- node[lab]{$\{\pvec_i\}_{i\in\Oenv}$} (140.6,-35.1);
\node[inner sep=0, anchor=west] at (141.2,-35.1) {\includegraphics[height=13.66mm]{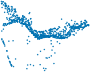}};
\draw[sup] (158.8,-35.1) -- (164.0,-35.1);
\fill[shadow] (164.9,-41.1) rectangle (176.7,-30.1);
\filldraw[fill=white, draw=cpOrange, line width=.7pt, rounded corners=1.6pt]
  (164.4,-40.6) rectangle (176.2,-29.6);
\node[align=center, inner sep=0] at (170.3,-35.1)
  {Chamfer\\[.5mm]{\normalsize$\mathcal{L}_{\mathrm{env}}$}};
\draw[flow] (170.3,-21.9) -- (170.3,-29.4);
\pgfresetboundingbox
\path[use as bounding box] (0,0) rectangle (177.8,-46.2);
\end{tikzpicture}
\caption{Overview of MORSE-Pose. The inference path combines multi-level RGB-D
features, adaptive keypoints, and learned category context to estimate pose,
size, and an observed-surface reconstruction. During training, pose and size
annotations define the envelope-selected support for object-aware reconstruction.
Blue highlights the proposed envelope operator, and orange denotes training-only
supervision.}
\label{fig:pipeline}
\end{figure*}

\section{MORSE-Pose}
\label{sec:method}

MORSE-Pose learns an object-centered representation from a segmented RGB-D
observation and a category label, built around a pose-and-size envelope that
converts standard category-level annotations into an object-aware
reconstruction target. Figure~\ref{fig:pipeline} summarizes the estimator and
the training-only supervision path.

\subsection{Problem Setup}
\label{sec:problem}

Let $\Pobs=\{\pvec_i\}_{i=1}^{N}$ with $\pvec_i\in\R^3$ denote the
back-projected point cloud of a segmented RGB-D crop, let $\mathbf{I}$ be its
RGB image, and let $c$ be its category. The target is the 9DoF pose
$(\Rgt,\tgt,\sgt)$, where $\Rgt\in SO(3)$, $\tgt\in\R^3$, and
$\sgt\in\R_{>0}^3$ gives the metric extent along the canonical axes. Following
NOCS~\cite{nocs}, an observed point is mapped to normalized object coordinates
as
\begin{equation}
  \qvec_i = \frac{1}{\lVert\sgt\rVert_2}\,
  \Rgt^{\!\top}(\pvec_i-\tgt),
  \label{eq:nocs}
\end{equation}
with normalized side lengths $e_a=s_a/\lVert\sgt\rVert_2$, a subscript
$a\in\{x,y,z\}$ denoting a vector's component along that canonical axis.

An adaptive-keypoint estimator predicts observed keypoints, their canonical
coordinates, the object pose and size, and a reconstructed point set
$\widehat{\Pobs}$ that keeps the sparse representation faithful to the measured
geometry. Because a segmented depth crop may also contain nearby support or
background points, the reconstruction target is restricted to an object-aware
support set $\Omega$:
\begin{equation}
  \mathcal{L}_{\mathrm{rec}}=
  \operatorname{CD}\big(\{\pvec_i\mid i\in\Omega\},\widehat{\Pobs}\big),
  \label{eq:rec}
\end{equation}
where $\operatorname{CD}$ is the symmetric Chamfer distance, the average of
the two directed mean nearest-neighbor distances.

A conventional surface-aware target constructs this set from the registered
instance mesh, represented as a set $\Mmesh$ of surface points expressed in
the same normalized object coordinates as $\qvec_i$:
\begin{equation}
 \Omesh=\left\{i:\min_{\mathbf{m}\in\Mmesh}
 \lVert\qvec_i-\mathbf{m}\rVert_2<\tau\right\},
 \label{eq:meshmask}
\end{equation}
where $\tau$ is an object-normalized tolerance. In MORSE-Pose, the measured
points, instance pose, and metric size determine the support set directly.
An RGB-D training observation with these labels can therefore supervise
object-aware reconstruction.

\subsection{Pose-and-Size Envelope}
\label{sec:envelope}

The metric-size label defines the normalized annotation box
\begin{equation}
 \mathcal{B}(\sgt)=\left\{\qvec:\lvert q_a\rvert\leq
 \frac{e_a}{2}\;\;\forall a\in\{x,y,z\}\right\},
 \label{eq:tightbox}
\end{equation}
with normalized side lengths $e_a$ from Sec.~\ref{sec:problem}. We dilate each
face of $\mathcal{B}(\sgt)$ by the tolerance $\tau$ of
Eq.~\eqref{eq:meshmask} and select
\begin{equation}
 \Oenv=\left\{i:\lvert q_{i,a}\rvert\leq
 \frac{e_a}{2}+\tau\;\;\forall a\in\{x,y,z\}\right\}.
 \label{eq:envmask}
\end{equation}
This axis-aligned envelope converts the pose, size, and measured point cloud
into an object-aware reconstruction target whose tolerance, normalized by the
object-size diagonal, keeps the same relative meaning across categories and
physical scales.

\textbf{Containment.}
Assume the annotation box bounds the registered surface,
$\Mmesh\subset\mathcal{B}(\sgt)$, so every $\mathbf{m}\in\Mmesh$ satisfies
$\lvert m_a\rvert\leq e_a/2$. For
any $i\in\Omesh$, there exists a surface point $\mathbf{m}$ with
$\lVert\qvec_i-\mathbf{m}\rVert_2<\tau$. Therefore
$\lvert q_{i,a}-m_a\rvert<\tau$ for every axis, and the triangle inequality
gives $\lvert q_{i,a}\rvert<e_a/2+\tau$. Hence
\begin{equation}
  \Omesh\subseteq\Oenv \qquad \text{for all }\tau\geq0.
  \label{eq:containment}
\end{equation}
The envelope is thus conservative with respect to the surface-distance target,
preserving its selected observations while extending support within the
object-scale neighborhood defined by the annotations. The inclusion can be
strict around concavities, hollow regions, and the dilated box boundary.

\begin{figure}[t]
\centering
\includegraphics[width=\linewidth]{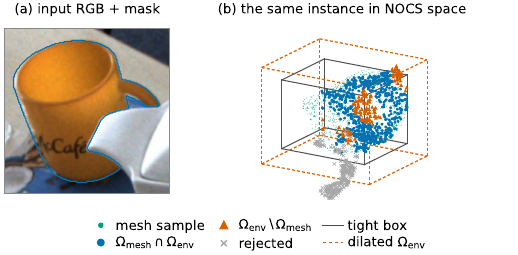}
\caption{Surface-distance and pose-and-size envelope selection on a HouseCat6D
observation. Panel (a) shows the RGB crop and instance mask. Panel (b) shows the
observed points in normalized object coordinates.}
\label{fig:maskexample}
\end{figure}

The construction is also independent of camera coordinates and metric units:
Eq.~\eqref{eq:nocs} removes any rigid transformation applied jointly to the
point cloud and pose, and $\qvec_i$ and $e_a$ share the size
normalization. Figure~\ref{fig:maskexample} illustrates the resulting support
on a HouseCat6D observation: both selectors retain the observed surface, and
the envelope adds points near the cup wall and interior while excluding the
support plane.

\subsection{Adaptive RGB-D Keypoint Representation}
\label{sec:keypoints}

\textbf{Multi-level point features.}
Let $\phi^{\mathrm{geo}}_i$ denote the geometric descriptor of $\pvec_i$, and
let
$\{\Phi_\ell(\mathbf{I})\}_{\ell\in\mathcal{S}}$ be feature maps from a set
$\mathcal{S}$ of stages of an image encoder~\cite{dinov2}, following
multi-scale feature practice~\cite{fpn,vitdesc}. With $u_i$ the pixel of
$\mathbf{I}$ from which
$\pvec_i$ is back-projected, and after aligning the feature maps to the crop,
the fused point feature is
\begin{equation}
  f_i=\Big[\phi^{\mathrm{geo}}_i,\;
  \psi_{\mathrm{rgb}}\!\big(
  \left[\Phi_\ell(\mathbf{I})(u_i)\right]_{\ell\in\mathcal{S}}
  \big)\Big],
  \label{eq:multilevel}
\end{equation}
where $\psi_{\mathrm{rgb}}$ projects the concatenated multi-level features to
a shared embedding, so $f_i$ combines local 3D geometry with multi-level
visual cues. Our models take $\mathcal{S}$
as four evenly spaced transformer blocks from the middle to the final layer of
a frozen DINOv2 ViT-B encoder, with $\psi_{\mathrm{rgb}}$ projecting to $128$
dimensions, the width of $\phi^{\mathrm{geo}}_i$, which the PointNet++
encoder~\cite{pointnet2} of AG-Pose produces.

\textbf{Adaptive keypoints and category context.}
Following AG-Pose~\cite{agpose}, shared slot queries produce, for keypoints
$j=1,\dots,K$, normalized assignments $w_{ji}$ over the observed points:
\begin{equation}
\begin{aligned}
  \mathbf{k}_j&=\sum_{i=1}^{N}w_{ji}\pvec_i, &
  \mathbf{g}_j&=\sum_{i=1}^{N}w_{ji}f_i,\\
  w_{ji}&\geq0, & \sum_{i=1}^{N} w_{ji}&=1.
\end{aligned}
  \label{eq:keypoints}
\end{equation}
Each keypoint is thus anchored to visible geometry while adapting to the
current instance, and local neighborhoods together with inter-keypoint
relations refine its feature~\cite{agpose}.

Prior methods encode category-level structure through deformable shape
priors~\cite{spd,sgpa,dpdn}, global category
embeddings~\cite{gcepose}, and curated feature dictionaries~\cite{cleanpose}.
MORSE-Pose learns a compact set of context prototypes $\mathbf{C}_c$
for each category. The keypoint features
$\mathbf{G}=[\mathbf{g}_1,\dots,\mathbf{g}_K]$ jointly attend to these
prototypes and to each other, receiving a residual update
\begin{equation}
  \widetilde{\mathbf{G}}=
  \mathbf{G}+\mathcal{A}(\mathbf{G},\mathbf{C}_c),
  \label{eq:context}
\end{equation}
where $\mathcal{A}$ denotes attention and feature
projection~\cite{transformer} and column $j$ of $\widetilde{\mathbf{G}}$ is
the updated feature $\widetilde{\mathbf{g}}_j$. The updated keypoints predict
canonical keypoint coordinates, rotation, translation, metric size, and the
reconstructed observed surface.
Each category owns $12$ prototype tokens of dimension $256$, the keypoint
feature width, and $\mathcal{A}$ stacks a few cross- and self-attention blocks
at that width.

\subsection{Training Objective}
\label{sec:objective}

The reconstruction branch is supervised by the envelope-selected observations,
\begin{equation}
 \mathcal{L}_{\mathrm{env}}=
 \operatorname{CD}\big(\{\pvec_i\mid i\in\Oenv\},\widehat{\Pobs}\big),
 \label{eq:envloss}
\end{equation}
and the complete objective is
\begin{equation}
 \mathcal{L}=\mathcal{L}_{\mathrm{pose}}+
 \lambda_{n}\mathcal{L}_{\mathrm{nocs}}+
 \lambda_{k}\mathcal{L}_{\mathrm{kpt}}+
 \lambda_{r}\mathcal{L}_{\mathrm{env}},
 \label{eq:full_loss}
\end{equation}
where the pose, canonical-coordinate, and keypoint terms follow the adaptive
keypoint formulation~\cite{agpose} and the scalar weights
$\lambda_{n},\lambda_{k},\lambda_{r}$ keep its values. Every target is defined
by the RGB-D observation, category, pose, metric size, or predicted keypoints,
and the envelope affects only reconstruction supervision, leaving all observed
points available to the feature and pose pathways. At inference, MORSE-Pose
receives an RGB-D crop and category label and directly predicts pose and size.

\section{Experiments}
\label{sec:experiments}

\subsection{Experimental Setup}
\label{sec:setup}

\textbf{Datasets and metrics.}
We evaluate on NOCS REAL275~\cite{nocs}, which contains six categories and uses
the synthetic CAMERA25 split for joint training, and on
HouseCat6D~\cite{housecat6d}, which spans ten household categories with diverse
scenes, viewpoints, materials, and occlusion. Instance masks follow the
protocol shared by recent methods: REAL275 inference uses the segmentation
results released by prior work~\cite{dpdn}, whereas HouseCat6D is trained and
evaluated with the ground-truth masks the benchmark provides. We report average
precision for 3D intersection over union and the joint thresholds \degtwo,
\degfive, \tendegtwo, and \tendegfive, with symmetric categories handled
according to each benchmark. REAL275 IoU$_{75}$ follows the corrected CATRE
protocol~\cite{catre}, consistent with recent comparisons.

\textbf{Training protocol.}
MORSE-Pose is trained from RGB-D observations, instance masks, category labels,
pose, and metric size. Unless noted, models use $N{=}1{,}024$ sampled points,
$K{=}96$ keypoints, and $\tau{=}0.1$. The benchmark models use $336{\times}336$ RGB crops and are trained for $55$ epochs on
REAL275 and $140$ epochs on HouseCat6D. Controlled target studies share a fixed
reference estimator, identical schedules, and three random seeds. The
reference configuration reaches
$53.3$ IoU$_{75}$ and $55.4$ \tendegtwo{}, against three-run means of $60.1$ and $61.2$ for the
benchmark configuration in Table~\ref{tab:housecat}.

\subsection{Main Results}
\label{sec:mainresults}

Tables~\ref{tab:real} and~\ref{tab:housecat} place MORSE-Pose among prior
methods on both benchmarks. All entries are AP in percent. IoU$_{75}^{\dagger}$
uses the corrected REAL275 protocol described in Sec.~\ref{sec:setup}. ``Geom.
sup.'' denotes direct use of CAD or mesh surfaces, sampled surface points, or
learned shape priors as training supervision. Label-supervised methods use
standard pose and size annotations. Bold and underlined values mark the best
and second-best results, ``--'' denotes an unreported entry, and
Table~\ref{tab:housecat} lists the methods that report HouseCat6D results.

\begin{table*}[!t]
\centering
\caption{Comparison with prior methods on REAL275. MORSE-Pose: mean of three training runs (std.\ ${\le}\,1.3$ AP).}
\label{tab:real}
\footnotesize
\renewcommand{\arraystretch}{0.95}
\setlength{\tabcolsep}{6.0pt}
\begin{tabular}{llcccccc}
\toprule
Method & Venue & Geom. sup. & IoU$_{75}^{\dagger}$ $\uparrow$ & \degtwo{} $\uparrow$ & \degfive{} $\uparrow$ & \tendegtwo{} $\uparrow$ & \tendegfive{} $\uparrow$ \\
\midrule
\multicolumn{8}{l}{\textbf{Non-keypoint-based}} \\
\addlinespace[1pt]
DPDN~\cite{dpdn} & ECCV\,'22 & Yes & 54.0 & 46.0 & 50.7 & 70.4 & 78.4 \\
HS-Pose~\cite{hspose} & CVPR\,'23 & No & 39.1 & 45.3 & 54.9 & 68.6 & 83.6 \\
VI-Net~\cite{vinet} & ICCV\,'23 & No & 48.3 & 50.0 & 57.6 & 70.8 & 82.1 \\
CatFormer~\cite{catformer} & AAAI\,'24 & Yes & -- & 47.7 & 53.7 & 69.0 & 79.5 \\
SecondPose~\cite{secondpose} & CVPR\,'24 & No & 49.7 & 56.2 & 63.6 & 74.7 & 86.0 \\
SpherePose~\cite{spherepose} & ICLR\,'25 & No & -- & 58.2 & 67.4 & 76.2 & 88.2 \\
\midrule
\multicolumn{8}{l}{\textbf{Keypoint-based}} \\
\addlinespace[1pt]
AG-Pose~\cite{agpose} & CVPR\,'24 & Yes & 61.3 & 57.0 & 64.6 & 75.1 & 84.7 \\
GCE-Pose~\cite{gcepose} & CVPR\,'25 & Yes & -- & 57.0 & 65.1 & 75.6 & 86.3 \\
SpotPose~\cite{spotpose} & CVPR\,'25 & Yes & -- & 59.7 & 64.8 & \underline{81.5} & 88.2 \\
MK-Pose~\cite{mkpose} & IROS\,'25 & Yes & -- & 60.8 & -- & 78.0 & 84.6 \\
CleanPose~\cite{cleanpose} & ICCV\,'25 & Yes & \underline{62.7} & \underline{61.7} & 67.6 & 78.3 & 86.3 \\
ComPose~\cite{compose} & CVPR\,'26 & Yes & -- & \textbf{62.1} & \textbf{68.0} & \textbf{81.8} & \textbf{89.2} \\
DecomPose~\cite{decompose} & ICML\,'26 & Yes & -- & 61.2 & \underline{67.9} & 79.2 & 87.5 \\
MemPose~\cite{mempose} & ECCV\,'26 & Yes & -- & 59.9 & 67.7 & 79.0 & -- \\
\addlinespace[1.5pt]
\oursrow
\textbf{MORSE-Pose} & This work & \textbf{No} & \textbf{64.2} & 59.1 & 66.0 & 79.6 & \underline{88.8} \\
\bottomrule
\end{tabular}
\end{table*}

\begin{table*}[!t]
\centering
\caption{Comparison with prior methods on HouseCat6D. MORSE-Pose: mean of three training runs (std.\ ${\le}\,1.6$ AP).}
\label{tab:housecat}
\footnotesize
\renewcommand{\arraystretch}{0.95}
\setlength{\tabcolsep}{4.6pt}
\begin{tabular}{llccccccccc}
\toprule
Method & Venue & Geom. sup. & IoU$_{25}$ $\uparrow$ & IoU$_{50}$ $\uparrow$ & IoU$_{75}$ $\uparrow$ & \degtwo{} $\uparrow$ & \degfive{} $\uparrow$ & \tendegtwo{} $\uparrow$ & \tendegfive{} $\uparrow$ \\
\midrule
\multicolumn{10}{l}{\textbf{Non-keypoint-based}} \\
\addlinespace[1pt]
FS-Net~\cite{fsnet} & CVPR\,'21 & Yes & 74.9 & 48.0 & -- & 3.3 & 4.2 & 17.1 & 21.6 \\
GPV-Pose~\cite{gpvpose} & CVPR\,'22 & No & 74.9 & 50.7 & -- & 3.5 & 4.6 & 17.8 & 22.7 \\
VI-Net~\cite{vinet} & ICCV\,'23 & No & 80.7 & 56.4 & -- & 8.4 & 10.3 & 20.5 & 29.1 \\
SecondPose~\cite{secondpose} & CVPR\,'24 & No & 83.7 & 66.1 & -- & 11.0 & 13.4 & 25.3 & 35.7 \\
SpherePose~\cite{spherepose} & ICLR\,'25 & No & 88.8 & 72.2 & -- & 19.3 & 25.9 & 40.9 & 55.3 \\
\midrule
\multicolumn{10}{l}{\textbf{Keypoint-based}} \\
\addlinespace[1pt]
AG-Pose~\cite{agpose} & CVPR\,'24 & Yes & 88.1 & 76.9 & 53.0 & 21.3 & 22.1 & 51.3 & 54.3 \\
GCE-Pose~\cite{gcepose} & CVPR\,'25 & Yes & -- & 79.2 & \textbf{60.6} & 24.8 & 25.7 & 55.4 & 58.4 \\
SpotPose~\cite{spotpose} & CVPR\,'25 & Yes & 89.1 & 77.0 & -- & 23.8 & 24.5 & 52.3 & 54.8 \\
CleanPose~\cite{cleanpose} & ICCV\,'25 & Yes & \underline{89.2} & 79.8 & 53.9 & 22.4 & 24.1 & 51.6 & 56.5 \\
ComPose~\cite{compose} & CVPR\,'26 & Yes & \textbf{90.3} & \underline{80.6} & -- & \underline{25.8} & \underline{27.6} & \underline{57.8} & \underline{61.5} \\
DecomPose~\cite{decompose} & ICML\,'26 & Yes & -- & 79.4 & 56.3 & 25.4 & 25.9 & 54.2 & 58.9 \\
MemPose~\cite{mempose} & ECCV\,'26 & Yes & -- & \textbf{81.5} & 56.4 & 23.1 & -- & 52.6 & 57.3 \\
\addlinespace[1.5pt]
\oursrow
\textbf{MORSE-Pose} & This work & \textbf{No} & 88.7 & 78.0 & \underline{60.1} & \textbf{30.6} & \textbf{31.1} & \textbf{61.2} & \textbf{63.2} \\
\bottomrule
\end{tabular}
\end{table*}

\begin{figure*}[t]
\centering
\begingroup
\footnotesize
\setlength{\tabcolsep}{0pt}
\renewcommand{\arraystretch}{1.0}
\newcommand{\qualrowwidth}{\dimexpr.8592\textwidth+11pt\relax}
\begin{tabular}{@{}
  >{\raggedright\arraybackslash}m{.1129\textwidth}
  @{\hspace{2pt}}
  *{5}{>{\centering\arraybackslash}m{.0716\textwidth}@{\hspace{.6pt}}}
  >{\centering\arraybackslash}m{.0716\textwidth}
  @{\hspace{5pt}}
  *{5}{>{\centering\arraybackslash}m{.0716\textwidth}@{\hspace{.6pt}}}
  >{\centering\arraybackslash}m{.0716\textwidth}
  @{}}
\toprule
\textbf{Method} &
\multicolumn{6}{c}{\textbf{NOCS-REAL275}} &
\multicolumn{6}{c}{\textbf{HouseCat6D}} \\
\cmidrule(lr){2-7}\cmidrule(lr){8-13}
& \emph{bottle} & \emph{bowl} & \emph{camera} & \emph{can} & \emph{laptop} & \emph{mug}
& \emph{shoe} & \emph{tube} & \emph{can} & \emph{box} & \emph{remote} & \emph{bottle} \\
\midrule
\nosup\textbf{GT} &
\multicolumn{12}{@{}c@{}}{\includegraphics[width=\qualrowwidth]{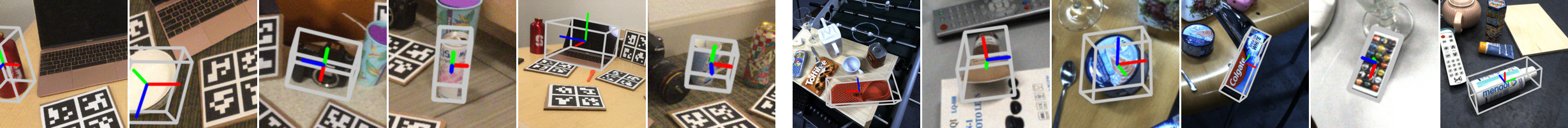}} \\[.6pt]
\labelsup VI-Net &
\multicolumn{12}{@{}c@{}}{\includegraphics[width=\qualrowwidth]{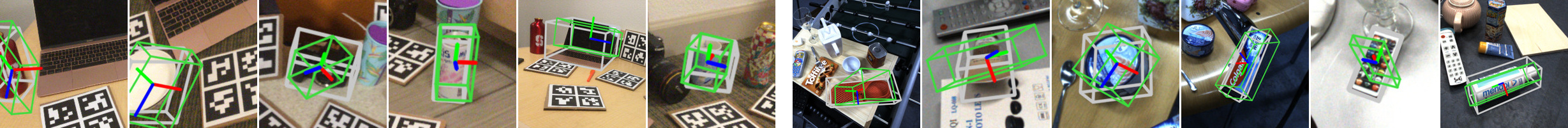}} \\[.6pt]
\meshsup AG-Pose &
\multicolumn{12}{@{}c@{}}{\includegraphics[width=\qualrowwidth]{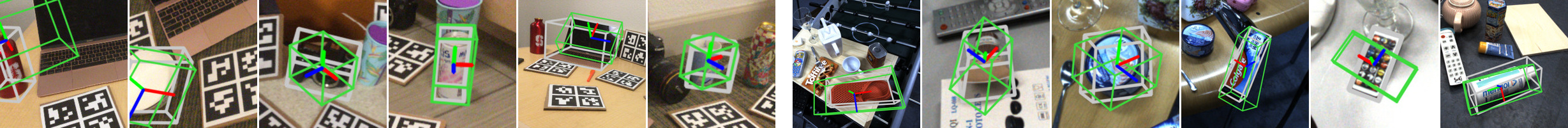}} \\[.6pt]
\meshsup CleanPose &
\multicolumn{12}{@{}c@{}}{\includegraphics[width=\qualrowwidth]{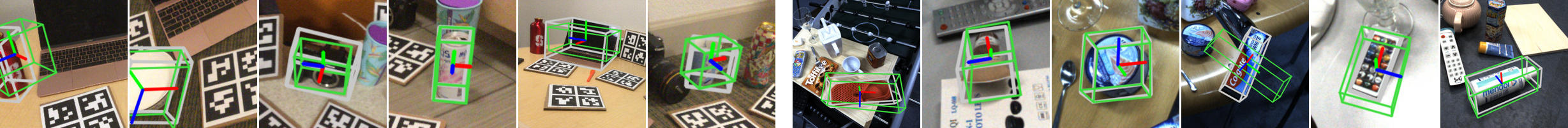}} \\[.6pt]
\labelsup\textbf{MORSE-Pose} &
\multicolumn{12}{@{}c@{}}{\includegraphics[width=\qualrowwidth]{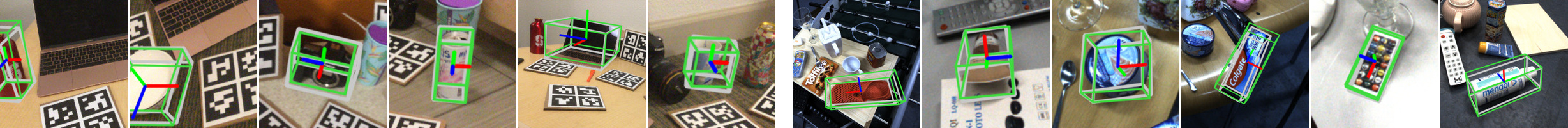}} \\
\bottomrule
\end{tabular}
\endgroup
\caption{Qualitative category-level pose predictions on NOCS-REAL275 and
HouseCat6D. Beside each method name, \meshsup{}denotes training with mesh or
shape-prior supervision and \labelsup{}denotes label-only supervision.}
\label{fig:qualitative}
\end{figure*}

Figure~\ref{fig:qualitative} visualizes predictions for all six NOCS-REAL275
categories and six representative HouseCat6D categories. The ground-truth row
shows the annotated box and object axes, and each method row overlays its
prediction. The comparison spans geometry-supervised and label-supervised
observation-driven methods.

On REAL275, MORSE-Pose achieves the highest corrected IoU$_{75}$ and the
second-highest \tendegfive{} AP among the methods in Table~\ref{tab:real}.
Its other scores are within $3.0$ AP of the best reported values.
Relative to the adaptive-keypoint baseline
AG-Pose~\cite{agpose}, it improves IoU$_{75}$ by $2.9$ AP and the four joint
metrics by $2.1/1.4/4.5/4.1$ AP. On HouseCat6D, MORSE-Pose has the highest
AP at all four joint pose thresholds in Table~\ref{tab:housecat}.
MemPose~\cite{mempose} scores $3.5$ AP higher at IoU$_{50}$. MORSE-Pose
scores $3.7$ AP higher at IoU$_{75}$ and $5.9$ AP higher at \tendegfive{}.

SpherePose~\cite{spherepose} has the highest joint pose scores among the
prior label-supervised methods in these tables. On REAL275, MORSE-Pose scores
$0.9$, $3.4$, and $0.6$ AP higher at \degtwo{}, \tendegtwo{}, and
\tendegfive{}, respectively. SpherePose scores $1.4$ AP higher at \degfive{}.
On HouseCat6D, MORSE-Pose improves on SpherePose by $11.3/5.2/20.3/7.9$ AP
at the four joint thresholds in table order. Figure~\ref{fig:qualitative}
shows stable pose and extent estimates across object geometries, with the most
visible variation in views where orientation is ambiguous. On REAL275, camera
and mug have the two lowest \degtwo{} scores. Holding translation tolerance
at $2$\,cm and increasing rotation tolerance to $10^\circ$ raises camera AP
from $3.4$ to $44.9$ and mug AP from $35.2$ to $92.8$.

\subsection{Envelope-to-Surface Correspondence}
\label{sec:maskanalysis}

Equation~\eqref{eq:containment} provides a one-sided relation between the two
support sets. Figure~\ref{fig:maskagreement} quantifies the additional envelope
support at $\tau{=}0.1$ over the HouseCat6D training observations, with
categories ordered by their mean smallest normalized extent
$e_{\min}{=}\min_a e_a$ and registered surfaces used only as an evaluation
reference. The dilated-to-tight box-volume ratio varies strongly with object
thinness, from $2.52$ for \emph{cup} to $20.06$ for \emph{cutlery} on
HouseCat6D, while the six REAL275 categories lie between $2.58$ and $3.02$. Point-level support is much more stable because
observed samples remain concentrated near object surfaces. Across HouseCat6D,
the envelope and surface-distance selector retain $91.3\%$ and $86.0\%$ of
sampled points, respectively. Envelope precision is $94.3\%$ overall and
$87.9\%$ for \emph{cutlery}, the category with the largest box-volume ratio.

In the containment audit, the envelope retained nearly all surface-selected
points. The few points it missed lay within numerical tolerance of the
dilated boundary. Roughly half of the
additional envelope points sit in the dilation shell outside the tight box, and
the remainder lie inside it around concavities or interiors. Their median
normalized surface distance is $0.11$, compared with $0.03$ for points selected
by both criteria. Selection is equally stable under coordinate changes: random
rigid camera transformations and metric-unit rescalings leave the decisions
unchanged beyond a $10^{-3}$ band around the envelope boundary.

\subsection{Controlled Target Substitution}
\label{sec:substitution}

\begin{figure}[t]
\centering
\includegraphics[width=\linewidth]{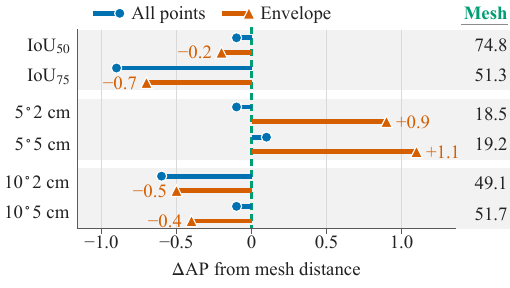}
\caption{Matched reconstruction-target substitution on HouseCat6D. Marker
position is the AP difference from the mesh-distance selector, whose absolute AP
is listed on the right. The estimator, schedule, and loss weights are shared,
with only reconstruction support changing (three-run means).}
\label{fig:substitution}
\end{figure}

Figure~\ref{fig:substitution} isolates reconstruction-target construction on
HouseCat6D. The estimator, schedule, and supervision weights are shared across
all selectors, only the support set for the reconstruction term changes, and
values are arithmetic means over three runs.

The mean absolute difference between the two object-aware targets is $0.6$ AP
across the six HouseCat6D metrics. The envelope scores higher at both
$5^\circ$ thresholds, and the surface-distance target scores higher on the
other four. Both targets also improve held-out reconstruction. Chamfer error
against the registered surface decreases from $3.29\times10^{-3}\,$m for the
all-points target to $2.58\times10^{-3}\,$m with the surface-distance target
and $2.65\times10^{-3}\,$m with the envelope, so the envelope captures $90\%$
of the reduction provided by the surface-derived target. In the matched
REAL275 study, the two targets agree within $0.1$ AP at all three IoU
thresholds. The AP differences (envelope minus surface distance) are
$+0.7/+1.0$ at the two $5^\circ$ thresholds and $-3.3/-2.0$ at the two
$10^\circ$ thresholds. REAL275's compact categories also
bring the two selectors closer at the point level, roughly halving the
disagreement seen on HouseCat6D.

\subsection{Robustness to Envelope Geometry and Boundary Leakage}
\label{sec:robustness}

\begin{figure}[t]
\centering
\includegraphics[width=\linewidth]{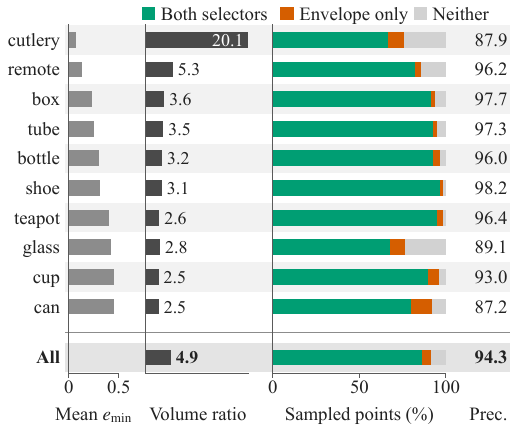}
\caption{Category-wise selector geometry and point-level agreement on
HouseCat6D at $\tau{=}0.1$. Categories are ordered by mean $e_{\min}$, and
\emph{All} denotes the aggregate. The first two panels are label-derived. In the
third, each bar groups sampled points by which selectors retain them,
using registered surfaces as the audit reference. The right column gives
envelope precision.}
\label{fig:maskagreement}
\end{figure}

\begin{figure}[t]
\centering
\includegraphics[width=\linewidth]{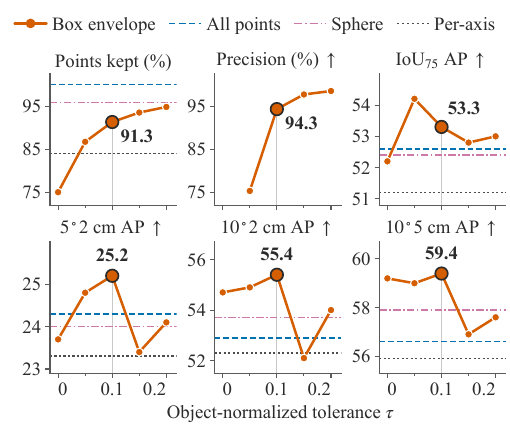}
\caption{Effect of object-normalized reconstruction-target tolerance and
selector shape on HouseCat6D. Each panel sweeps the box envelope over
$\tau\in[0,0.2]$, and the ringed marker with its label gives the $\tau{=}0.1$
operating point. Horizontal lines are the categorical reference selectors
evaluated with the same configuration. Values are percentages, with task
metrics reported as AP. Precision is shown only where a matched surface audit
was performed. The all-points and $\tau{=}0.1$ entries are three-run means.}
\label{fig:tolerance}
\end{figure}

Figure~\ref{fig:tolerance} evaluates tolerance and selector shape with a common
HouseCat6D model configuration, reporting the point-level statistics of
Fig.~\ref{fig:maskagreement} alongside task AP. Across $\tau\in[0,0.2]$, point
retention ranges from $75.0\%$ to $94.8\%$ and envelope precision rises to
$98.5\%$, while the largest change from the $\tau{=}0.1$ operating point at
\degtwo{} is $1.9$ AP. Alternative selector
shapes remain within $3.5$ AP of the operating point on all reported metrics,
with the full axis-aligned extent giving the most balanced result. This broad
plateau supports one object-relative tolerance across categories, and
$\tau{=}0.1$ is used on both benchmarks. At the two $10^\circ$ thresholds,
the $\tau{=}0.1$ envelope scores $2.5$ and $2.8$ AP higher than the all-points
target. The object-scaled selectors yield similar AP across substantially
different retained volumes. This pattern suggests that much of the gain comes
from selecting a reconstruction region at the scale of the object.

\begin{figure}[t]
\centering
\includegraphics[width=\linewidth]{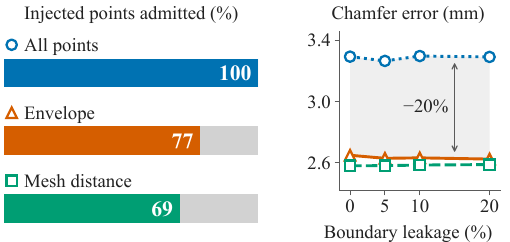}
\caption{Robustness to synthetic segmentation-boundary leakage on HouseCat6D
(three-run means).}
\label{fig:boundary}
\end{figure}

Boundary leakage probes the same construction under predicted-mask noise. We
augment each object sample with ring points just outside the instance boundary,
retrain at three contamination levels, and average each condition over three
runs. Figure~\ref{fig:boundary} reports the fraction of injected points
admitted to the reconstruction target and clean-scene Chamfer error against
the registered surface.

The envelope admits approximately $77\%$ of injected points, compared with
$69\%$ for the surface-distance target and $100\%$ without filtering, and this
rate remains stable across leakage levels. Its clean-scene reconstruction error
moves by at most $2.5\times10^{-5}\,$m and stays about $20\%$ below the
all-points target. Averaged over the leakage range, IoU$_{75}$ decreases by
$1.9$ AP for the envelope, against $2.9$ for the surface-distance target and
$2.8$ for the all-points target, and all \degtwo{} changes stay within $1.0$
AP.

\subsection{Cross-View Canonical Consistency}
\label{sec:crossview}

Cross-view consistency measures the stability of the learned canonical
representation: we pair observations of the same physical instance and align
their predictions through the annotated poses.
Table~\ref{tab:crossview} uses $300$ view pairs from $50$ instances, balanced
across the ten HouseCat6D categories. Keypoint repeatability, surface
residual, and canonical-coordinate error are in $10^{-3}$ units of the size
diagonal, \degtwo{} is evaluated on the paired views, and all entries are
three-run means.

\begin{table}[t]
\centering
\caption{Cross-view canonical consistency on HouseCat6D.}
\label{tab:crossview}
\footnotesize
\renewcommand{\arraystretch}{1.0}
\setlength{\tabcolsep}{3.6pt}
\begin{tabular}{@{}lcccc@{}}
\toprule
Selector & Repeat.\ $\downarrow$ & Surf.\ dist.\ $\downarrow$ & NOCS RMSE $\downarrow$ & \degtwo{} $\uparrow$ \\
\midrule
\multicolumn{5}{@{}l}{\textbf{All 300 pairs}} \\
\addlinespace[1pt]
All points    & 132.1 & 34.7 & 84.2 & 31.3 \\
Mesh distance & 127.7 & 33.4 & 79.2 & 29.4 \\
Envelope      & 127.5 & 33.3 & 81.7 & 30.2 \\
\midrule
\multicolumn{5}{@{}l}{\textbf{Low-overlap pairs}} \\
\addlinespace[1pt]
All points    & 214.6 & 37.6 & 92.8 & 27.8 \\
Mesh distance & 206.9 & 36.1 & 88.0 & 26.0 \\
Envelope      & 207.0 & 36.4 & 90.6 & 25.9 \\
\bottomrule
\end{tabular}
\end{table}

Both object-aware targets reduce keypoint drift, surface residual, and
canonical-coordinate error relative to the all-points target. Their geometric
scores also remain close on low-overlap pairs, where keypoint drift increases
by roughly $1.6\times$. The all-points target gives the highest paired-view
\degtwo{} score. This pose threshold and the three geometric measures capture
complementary properties of the predictions across views.

\subsection{Ablation Study}
\label{sec:ablation}

Table~\ref{tab:componentablation} varies one component at a time from the
full model in Table~\ref{tab:housecat}. Using the multi-level features of
Eq.~\eqref{eq:multilevel} in place of the final encoder stage alone adds
$5.2$ AP at both \degtwo{} and \tendegtwo{}. The category context prototypes
of Eq.~\eqref{eq:context} add $1.0$ and $1.7$ AP on these metrics, and a
ViT-B backbone improves on ViT-S by $4.7$ and $4.8$ AP. AG-Pose's
non-register ViT-B/14 control reaches $22.4/23.9/55.8/60.3$ AP.
MORSE-Pose scores $8.2/7.2/5.4/2.9$ AP higher. Of the two representation
components, multi-level features contribute the larger gains. The context
prototypes have their largest effect at the $10^\circ$ thresholds.

\begin{table}[t]
\centering
\caption{Ablation on HouseCat6D. Indented rows vary one element of
MORSE-Pose (\meshsup{}mesh supervision, \labelsup{}label-only training).}
\label{tab:componentablation}
\footnotesize
\renewcommand{\arraystretch}{1.0}
\setlength{\tabcolsep}{3.0pt}
\begin{tabular}{@{}lcccc@{}}
\toprule
Configuration & \degtwo{} & \degfive{} & \tendegtwo{} & \tendegfive{} \\
\midrule
\labelsup\textbf{MORSE-Pose} (full) & 30.6 & 31.1 & 61.2 & 63.2 \\
\ \ w/o multi-level $\Phi_\ell$   & 25.4 & 26.7 & 56.0 & 60.0 \\
\ \ w/o context $\mathbf{C}_c$    & 29.6 & 30.5 & 59.5 & 61.7 \\
\ \ with ViT-S backbone           & 25.9 & 26.6 & 56.4 & 58.6 \\
\midrule
\meshsup AG-Pose (ViT-S) & 21.3 & 22.1 & 51.3 & 54.3 \\
\meshsup AG-Pose (ViT-B) & 22.4 & 23.9 & 55.8 & 60.3 \\
\bottomrule
\end{tabular}
\end{table}

Under a shortened one-cycle schedule of $50$ epochs, the four encoder stages
used for the benchmarks yield $23.9$ AP at \degtwo{}. Four earlier or later
stages yield $21.2$ and $23.3$ AP, respectively. Six stages yield $22.9$ AP,
and the final stage alone yields $18.4$ AP. A two-stage variant gains $1.9$
AP at \degtwo{}. The larger differences at IoU$_{75}$ and the $10^\circ$
thresholds favor the four stages spread across the encoder depth.

\subsection{Model Size and Inference Cost}
\label{sec:efficiency}

Table~\ref{tab:efficiency} reports HouseCat6D joint pose AP and efficiency
under a common frozen DINOv2 ViT-B/14 setting~\cite{dinov2}. AG-Pose and
CleanPose follow their released recipes with ViT-B/14 backbones. Costs use one
RTX A6000, $1{,}024$ points, and $300$ forward passes after warm-up.
MORSE-Pose uses $101.9$M parameters with multi-level fusion and learned
category prototypes. AG-Pose uses $98.1$M, and CleanPose uses $101.6$M with
its inference-time feature dictionary. Inference takes $28.8$ ms for
MORSE-Pose, $23.4$ ms for AG-Pose, and $31.0$ ms for CleanPose.
MORSE-Pose projects features before upsampling and uses $551$ MiB of peak
memory, the lowest of the three models.

\begin{table}[t]
\centering
\caption{HouseCat6D ViT-B/14 pose AP/efficiency (M/MiB/ms; batch 1).}
\label{tab:efficiency}
\scriptsize
\renewcommand{\arraystretch}{1.0}
\setlength{\tabcolsep}{1.8pt}
\resizebox{\columnwidth}{!}{%
\begin{tabular}{@{}lccccccc@{}}
\toprule
Method & \degtwo{} & \degfive{} & \tendegtwo{} & \tendegfive{}
& Param. & Mem. & Time \\
\midrule
AG-Pose~\cite{agpose}       & 22.4 & 23.9 & 55.8 & 60.3 & 98.1  & 708 & \textbf{23.4} \\
CleanPose~\cite{cleanpose}  & 24.5 & 25.9 & 54.5 & 58.5 & 101.6 & 722 & 31.0 \\
\textbf{MORSE-Pose}         & \textbf{30.6} & \textbf{31.1} & \textbf{61.2} & \textbf{63.2} & 101.9 & \textbf{551} & 28.8 \\
\bottomrule
\end{tabular}
}
\end{table}

\FloatBarrier

\balance
\section{Conclusion}
\label{sec:conclusion}

We presented MORSE-Pose, which derives object-aware reconstruction support
from pose and metric-size annotations through a canonical envelope with
containment and invariance properties. Experiments on REAL275 and HouseCat6D
show stable reconstruction support, strong pose estimation, and consistent
behavior under perturbations. The pose and size labels used to train the
estimator can also supervise reconstruction of the observed geometry during
keypoint learning.

\bibliographystyle{IEEEtran}
\bibliography{refs}

@inproceedings{nocs,
  author    = {Wang, He and Sridhar, Srinath and Huang, Jingwei and others},
  title     = {Normalized Object Coordinate Space for Category-Level 6D Object Pose and Size Estimation},
  booktitle = {Proc. IEEE/CVF Conf. Comput. Vis. Pattern Recognit. (CVPR)},
  year      = {2019}
}

@inproceedings{housecat6d,
  author    = {Jung, HyunJun and Wu, Shun-Cheng and Ruhkamp, Patrick and others},
  title     = {{HouseCat6D} - A Large-Scale Multi-Modal Category Level {6D} Object Perception Dataset with Household Objects in Realistic Scenarios},
  booktitle = {Proc. IEEE/CVF Conf. Comput. Vis. Pattern Recognit. (CVPR)},
  pages     = {22498--22508},
  doi       = {10.1109/CVPR52733.2024.02123},
  year      = {2024}
}

@inproceedings{agpose,
  author    = {Lin, Xiao and Yang, Wenfei and Gao, Yuan and Zhang, Tianzhu},
  title     = {Instance-Adaptive and Geometric-Aware Keypoint Learning for Category-Level 6D Object Pose Estimation},
  booktitle = {Proc. IEEE/CVF Conf. Comput. Vis. Pattern Recognit. (CVPR)},
  pages     = {21040--21049},
  year      = {2024}
}

@inproceedings{cleanpose,
  author    = {Lin, Xiao and Peng, Yun and Wang, Liuyi and others},
  title     = {{CleanPose}: Category-Level Object Pose Estimation via Causal Learning and Knowledge Distillation},
  booktitle = {Proc. IEEE/CVF Int. Conf. Comput. Vis. (ICCV)},
  pages     = {5990--6000},
  doi       = {10.1109/ICCV51701.2025.00566},
  year      = {2025}
}

@inproceedings{gcepose,
  author    = {Li, Weihang and Xu, Hongli and Huang, Junwen and others},
  title     = {{GCE-Pose}: Global Context Enhancement for Category-level Object Pose Estimation},
  booktitle = {Proc. IEEE/CVF Conf. Comput. Vis. Pattern Recognit. (CVPR)},
  pages     = {27154--27165},
  doi       = {10.1109/CVPR52734.2025.02529},
  year      = {2025}
}

@inproceedings{compose,
  author    = {Ren, Huan and Chen, Yihan and Wang, Chuxin and others},
  title     = {{ComPose}: A Unified Completion-Pose Framework for Robust Category-Level Object Pose Estimation},
  booktitle = {Proc. IEEE/CVF Conf. Comput. Vis. Pattern Recognit. (CVPR)},
  pages     = {14315--14324},
  year      = {2026}
}

@inproceedings{decompose,
  author    = {Gao, Yifan and Zou, Lu and Huang, Zhangjin and others},
  title     = {{DecomPose}: Disentangling Cross-Category Optimization Contention for Category-Level {6D} Object Pose Estimation},
  booktitle = {Proc. Int. Conf. Mach. Learn. (ICML)},
  volume    = {306},
  series    = {Proc. Mach. Learn. Res.},
  publisher = {PMLR},
  year      = {2026}
}

@inproceedings{mempose,
  author    = {Lin, Xiao and Zhu, Minghao and Peng, Yun and others},
  title     = {{MemPose}: Category-Level Object Pose Estimation with Memory},
  booktitle = {Proc. Eur. Conf. Comput. Vis. (ECCV)},
  year      = {2026}
}

@inproceedings{spd,
  author    = {Tian, Meng and Ang, Marcelo H. and Lee, Gim Hee},
  title     = {Shape Prior Deformation for Categorical 6D Object Pose and Size Estimation},
  booktitle = {Proc. Eur. Conf. Comput. Vis. (ECCV)},
  pages     = {530--546},
  year      = {2020}
}

@inproceedings{dpdn,
  title={Category-level 6d object pose and size estimation using self-supervised deep prior deformation networks},
  author={Lin, Jiehong and Wei, Zewei and Ding, Changxing and Jia, Kui},
  booktitle={European Conference on Computer Vision},
  pages={19--34},
  year={2022},
  organization={Springer}
}

@inproceedings{catformer,
  author    = {Yu, Sheng and Zhai, Di-Hua and Xia, Yuanqing},
  title     = {{CatFormer}: Category-Level {6D} Object Pose Estimation with Transformer},
  booktitle = {Proc. AAAI Conf. Artif. Intell. (AAAI)},
  volume    = {38},
  number    = {7},
  pages     = {6808--6816},
  doi       = {10.1609/aaai.v38i7.28505},
  year      = {2024}
}

@inproceedings{fsnet,
  author    = {Chen, Wei and Jia, Xi and Chang, Hyung Jin and others},
  title     = {FS-Net: Fast Shape-Based Network for Category-Level 6D Object Pose Estimation with Decoupled Rotation Mechanism},
  booktitle = {Proc. IEEE/CVF Conf. Comput. Vis. Pattern Recognit. (CVPR)},
  pages     = {1581--1590},
  year      = {2021}
}

@inproceedings{gpvpose,
  author    = {Di, Yan and Zhang, Ruida and Lou, Zhiqiang and others},
  title     = {GPV-Pose: Category-Level Object Pose Estimation via Geometry-Guided Point-Wise Voting},
  booktitle = {Proc. IEEE/CVF Conf. Comput. Vis. Pattern Recognit. (CVPR)},
  pages     = {6781--6791},
  year      = {2022}
}

@inproceedings{hspose,
  author    = {Zheng, Linfang and Wang, Chen and Sun, Yinghan and others},
  title     = {HS-Pose: Hybrid Scope Feature Extraction for Category-Level Object Pose Estimation},
  booktitle = {Proc. IEEE/CVF Conf. Comput. Vis. Pattern Recognit. (CVPR)},
  pages     = {17163--17173},
  year      = {2023}
}

@inproceedings{vinet,
  author    = {Lin, Jiehong and Wei, Zewen and Zhang, Yuwei and others},
  title     = {VI-Net: Boosting Category-Level 6D Object Pose Estimation via Learning Decoupled Rotations on the Spherical Representations},
  booktitle = {Proc. IEEE/CVF Int. Conf. Comput. Vis. (ICCV)},
  pages     = {14001--14011},
  year      = {2023}
}

@inproceedings{istnet,
  author    = {Liu, Jianhui and Chen, Yukang and Ye, Xiaoqing and others},
  title     = {IST-Net: Prior-Free Category-Level Pose Estimation with Implicit Space Transformation},
  booktitle = {Proc. IEEE/CVF Int. Conf. Comput. Vis. (ICCV)},
  pages     = {13978--13988},
  year      = {2023}
}

@inproceedings{query6dof,
  author    = {Wang, Ruiqi and Wang, Xian and Li, Tinghui and others},
  title     = {Query6DoF: Learning Sparse Queries as Implicit Shape Prior for Category-Level 6DoF Pose Estimation},
  booktitle = {Proc. IEEE/CVF Int. Conf. Comput. Vis. (ICCV)},
  pages     = {14055--14064},
  year      = {2023}
}

@inproceedings{secondpose,
  author    = {Chen, Yamei and Di, Yan and Zhai, Guangyao and others},
  title     = {SecondPose: SE(3)-Consistent Dual-Stream Feature Fusion for Category-Level Pose Estimation},
  booktitle = {Proc. IEEE/CVF Conf. Comput. Vis. Pattern Recognit. (CVPR)},
  year      = {2024}
}

@article{dinov2,
  author  = {Oquab, Maxime and Darcet, Timoth{\'e}e and Moutakanni, Th{\'e}o and Vo, Huy and Szafraniec, Marc and Khalidov, Vasil and Fernandez, Pierre and Haziza, Daniel and Massa, Francisco and El-Nouby, Alaaeldin and others},
  title   = {DINOv2: Learning Robust Visual Features without Supervision},
  journal = {Trans. Mach. Learn. Res.},
  year    = {2024}
}

@inproceedings{catre,
  author    = {Liu, Xingyu and Wang, Gu and Li, Yi and others},
  title     = {CATRE: Iterative Point Clouds Alignment for Category-Level Object Pose Refinement},
  booktitle = {Proc. Eur. Conf. Comput. Vis. (ECCV)},
  pages     = {499--516},
  year      = {2022}
}

@inproceedings{foundationpose,
  author    = {Wen, Bowen and Yang, Wei and Kautz, Jan and others},
  title     = {{FoundationPose}: Unified {6D} Pose Estimation and Tracking of Novel Objects},
  booktitle = {Proc. IEEE/CVF Conf. Comput. Vis. Pattern Recognit. (CVPR)},
  pages     = {17868--17879},
  doi       = {10.1109/CVPR52733.2024.01692},
  year      = {2024}
}

@inproceedings{onepose,
  author    = {Sun, Jiaming and Wang, Zihao and Zhang, Siyu and others},
  title     = {{OnePose}: One-Shot Object Pose Estimation Without {CAD} Models},
  booktitle = {Proc. IEEE/CVF Conf. Comput. Vis. Pattern Recognit. (CVPR)},
  pages     = {6825--6834},
  doi       = {10.1109/CVPR52688.2022.00670},
  year      = {2022}
}

@inproceedings{sgpa,
  author    = {Chen, Kai and Dou, Qi},
  title     = {SGPA: Structure-Guided Prior Adaptation for Category-Level 6D Object Pose Estimation},
  booktitle = {Proc. IEEE/CVF Int. Conf. Comput. Vis. (ICCV)},
  pages     = {2773--2782},
  year      = {2021}
}

@inproceedings{spherepose,
  author    = {Ren, Huan and Yang, Wenfei and Liu, Xiang and others},
  title     = {Learning Shape-Independent Transformation via Spherical Representations for Category-Level Object Pose Estimation},
  booktitle = {Int. Conf. Learn. Represent. (ICLR)},
  year      = {2025}
}

@inproceedings{spotpose,
  author    = {Ren, Huan and Yang, Wenfei and Zhang, Shifeng and others},
  title     = {Rethinking Correspondence-Based Category-Level Object Pose Estimation},
  booktitle = {Proc. IEEE/CVF Conf. Comput. Vis. Pattern Recognit. (CVPR)},
  pages     = {1170--1179},
  year      = {2025}
}

@inproceedings{mkpose,
  author    = {Yang, Yifan and Song, Peili and Lan, Enfan and others},
  title     = {{MK-Pose}: Category-Level Object Pose Estimation via Multimodal-Based Keypoint Learning},
  booktitle = {Proc. IEEE/RSJ Int. Conf. Intell. Robots Syst. (IROS)},
  pages     = {4232--4239},
  doi       = {10.1109/IROS60139.2025.11247325},
  year      = {2025}
}

@inproceedings{posecnn,
  author    = {Xiang, Yu and Schmidt, Tanner and Narayanan, Venkatraman and others},
  title     = {{PoseCNN}: A Convolutional Neural Network for 6D Object Pose Estimation in Cluttered Scenes},
  booktitle = {Proc. Robot.: Sci. Syst. (RSS)},
  year      = {2018}
}

@inproceedings{densefusion,
  author    = {Wang, Chen and Xu, Danfei and Zhu, Yuke and others},
  title     = {{DenseFusion}: 6D Object Pose Estimation by Iterative Dense Fusion},
  booktitle = {Proc. IEEE/CVF Conf. Comput. Vis. Pattern Recognit. (CVPR)},
  pages     = {3343--3352},
  year      = {2019}
}

@inproceedings{transformer,
  author    = {Vaswani, Ashish and Shazeer, Noam and Parmar, Niki and others},
  title     = {Attention Is All You Need},
  booktitle = {Proc. Adv. Neural Inf. Process. Syst. (NeurIPS)},
  year      = {2017}
}

@inproceedings{fpn,
  author    = {Lin, Tsung-Yi and Doll{\'a}r, Piotr and Girshick, Ross and others},
  title     = {Feature Pyramid Networks for Object Detection},
  booktitle = {Proc. IEEE Conf. Comput. Vis. Pattern Recognit. (CVPR)},
  pages     = {2117--2125},
  year      = {2017}
}

@inproceedings{vitdesc,
  author    = {Amir, Shir and Gandelsman, Yossi and Bagon, Shai and others},
  title     = {Deep {ViT} Features as Dense Visual Descriptors},
  booktitle = {Proc. Eur. Conf. Comput. Vis. Workshops (ECCVW)},
  year      = {2022}
}

@inproceedings{pointnet2,
  author    = {Qi, Charles Ruizhongtai and Yi, Li and Su, Hao and others},
  title     = {{PointNet++}: Deep Hierarchical Feature Learning on Point Sets in a Metric Space},
  booktitle = {Proc. Adv. Neural Inf. Process. Syst. (NeurIPS)},
  year      = {2017}
}

\end{document}